\documentclass{article} 
\usepackage{lmrl2025_workshop,times}

\usepackage{amsmath,amsfonts,bm}

\def\eqref#1{equation~\ref{#1}}

\def\1{\bm{1}}

\DeclareMathAlphabet{\mathsfit}{\encodingdefault}{\sfdefault}{m}{sl}
\SetMathAlphabet{\mathsfit}{bold}{\encodingdefault}{\sfdefault}{bx}{n}

\usepackage{hyperref}
\usepackage{url}
\usepackage{graphicx}
\usepackage{cleveref}
\usepackage{booktabs}
\usepackage{subcaption}
\usepackage{multirow}
\usepackage{svg}

\usepackage{tikz}
\usetikzlibrary{shapes.geometric, arrows, positioning}

\tikzstyle{startstop} = [rectangle, rounded corners, minimum width=3cm, minimum height=1cm,text centered, draw=black, fill=red!30]
\tikzstyle{process} = [rectangle, rounded corners, minimum width=3cm, minimum height=1cm, align=center, draw=black, fill=blue!20, inner sep=8pt]
\tikzstyle{arrow} = [thick,->,>=stealth]
\tikzstyle{data} = [rectangle, minimum width=3cm, minimum height=1cm, text centered, draw=black, dashed]

\title{Synthetic Data for \\Blood Vessel Network Extraction}

\lmrlfinalcopy

\author{Joël Mathys, Andreas Plesner, Jorel Elmiger \& Roger Wattenhofer\\
ETH Zurich, Zurich, Switzerland \\
\texttt{\{jmathys,aplesner,wattenhofer\}@ethz.ch}
}

\begin{document}

\maketitle

\begin{abstract}
Blood vessel networks in the brain play a crucial role in stroke research, where understanding their topology is essential for analyzing blood flow dynamics. However, extracting detailed topological vessel network information from microscopy data remains a significant challenge, mainly due to the scarcity of labeled training data and the need for high topological accuracy. 
This work combines synthetic data generation with deep learning to automatically extract vessel networks as graphs from volumetric microscopy data.
To combat data scarcity, we introduce a comprehensive pipeline for generating large-scale synthetic datasets that mirror the characteristics of real vessel networks. Our three-stage approach progresses from abstract graph generation through vessel mask creation to realistic medical image synthesis, incorporating biological constraints and imaging artifacts at each stage. 
Using this synthetic data, we develop a two-stage deep learning pipeline of 3D U-Net-based models for node detection and edge prediction.
Fine-tuning on real microscopy data shows promising adaptation, improving edge prediction F1 scores from 0.496 to 0.626 by training on merely 5 manually labeled samples. These results suggest that automated vessel network extraction is becoming practically feasible, opening new possibilities for large-scale vascular analysis in stroke research.

\end{abstract}

\section{Introduction}
Understanding the topology of brain vessel networks is crucial for analyzing blood flow dynamics, particularly in stroke research, where vascular disruptions can have severe consequences \citep{shih2013smallest}. These networks can be modeled as graphs, enabling detailed flow analysis. However, accurately extracting vessel structures from high-resolution 3D microscopy data remains a significant challenge \citep{Schmid2017CerebralBF}.
Imaging noise, artifacts, and complex vessel junctions frequently lead to segmentation errors, disrupting downstream analyses.

Deep learning methods offer a promising approach to automate vessel network extraction, but the scarcity of labeled training data constrains their effectiveness. Annotating 3D vessel structures is an expert-intensive and time-consuming process, making creating the large-scale datasets required for robust model training challenging. This data bottleneck limits deep learning solutions' applicability, scalability, and generalization \citep{tetteh2020deepvesselnet,wittmann2024vesselfmfoundationmodeluniversal}.

We propose a synthetic data generation pipeline to address this challenge and improve vessel graph extraction in microscopy images. Our approach simulates realistic vessel networks based on physiological principles, generates volumetric imaging data with domain-specific artifacts, and produces large-scale synthetic datasets for training models. By leveraging synthetic data, we enhance model robustness, particularly in detecting complex junctions, and reduce reliance on extensive manual annotations. Fine-tuning on a minimal set of real microscopy samples further demonstrates the potential for synthetic data to bridge the gap between limited real-world data and scalable AI solutions.

This work highlights how synthetic data can mitigate critical data access limitations in biomedical AI, aligning with the broader goal of expanding machine learning applications in data-scarce domains. Our findings contribute to the discussion on the benefits and challenges of synthetic data for model training, supporting the development of more generalizable and scalable AI systems in biomedical research.


\begin{figure}[t]
    \centering
    \includegraphics[width=0.8\linewidth]{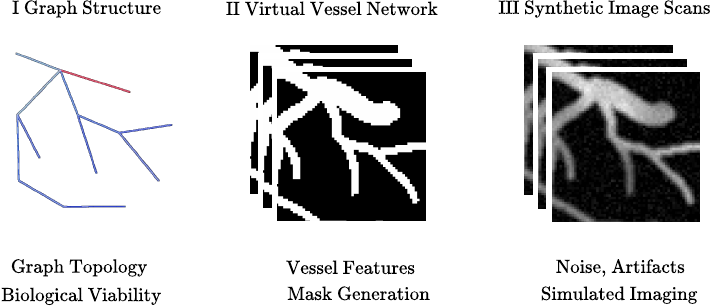}
    \vspace{-2mm}
    \caption{The synthetic data generation is split into three phases. First, the ground truth graph structure, including all node and edge placements, is generated while considering biological constraints. Then, Vessel features, such as the outline and curves, are generated. Finally, we add noise, artifacts, and distortions during the simulated imaging to derive a synthetic image of a blood vessel network.}
    \label{fig: data generation pipeline}
    \vspace{-3mm}
\end{figure}

\section{Synthetic Data Generation}


Blood vessel imaging data typically consists of 3D volumetric scans where vessels appear as tube-like structures against a dark background. The diameters of these structures can vary drastically, ranging from large arteries to fine capillaries. Moreover, the data frequently contains imaging artifacts, such as intensity variations along vessels and noise patterns that can vary with imaging depth and conditions. One of the main challenges besides the scan quality is the lack of labeled annotated ground truth data. We introduce a fully parametrized synthetic data generation pipeline to address the data scarcity for vessel network extraction (\Cref{fig: data generation pipeline}). Our approach enables the creation of large-scale training datasets while maintaining control over key characteristics of the generated samples, which can provide the ground truth of the topological graph structure. 

Our data generation pipeline consists of three main stages. First, we generate the underlying graph structure representing the vessel network. Here, we incorporate biological constraints such as realistic branching patterns and vessel diameter relationships following Murray's law \cite{Murray1926}. The second stage converts these abstract graphs into volumetric vessel masks to recreate a fully virtual model of the underlying vessel network. We use Bézier curves to create natural-looking vessel segments throughout the network. Finally, we simulate realistic imaging conditions by adding appropriate noise patterns and imaging artifacts that match those observed in real data.

\begin{table}[b]
    \centering
    \caption{Overview of the used datasets. We have access to 10 real data samples that are annotated to have the network topology. We generate two synthetic datasets, both of which contain at least 100k samples. The homogenous dataset contains samples from the configuration that resemble the real data the closest, whereas the Varied dataset consists of a diverse range of parameter configurations. }
    \label{dataset-comparison}
    \begin{tabular}{llllll}
        \multicolumn{1}{c}{\bf Dataset} & \multicolumn{1}{c}{\bf Samples} & \multicolumn{1}{c}{\bf Method} &  \multicolumn{1}{c}{\bf Size} & \multicolumn{1}{c}{\bf Source} \\
        \hline 
        Labeled Scans & 10 & TPLSM &  X GB & Real \\
        Homogeneous (ours) & 100k, 1M & Single Parameterset& X GB & Synthetic \\
        Varied (ours)& 1.08M & Diverse Parameterset & X GB & Synthetic\\
    \end{tabular}
    \label{tab: datasets}
\end{table}

\begin{figure}[t]
    \centering
    \begin{minipage}{0.45\textwidth}
        \centering
        \includegraphics[width=\textwidth]{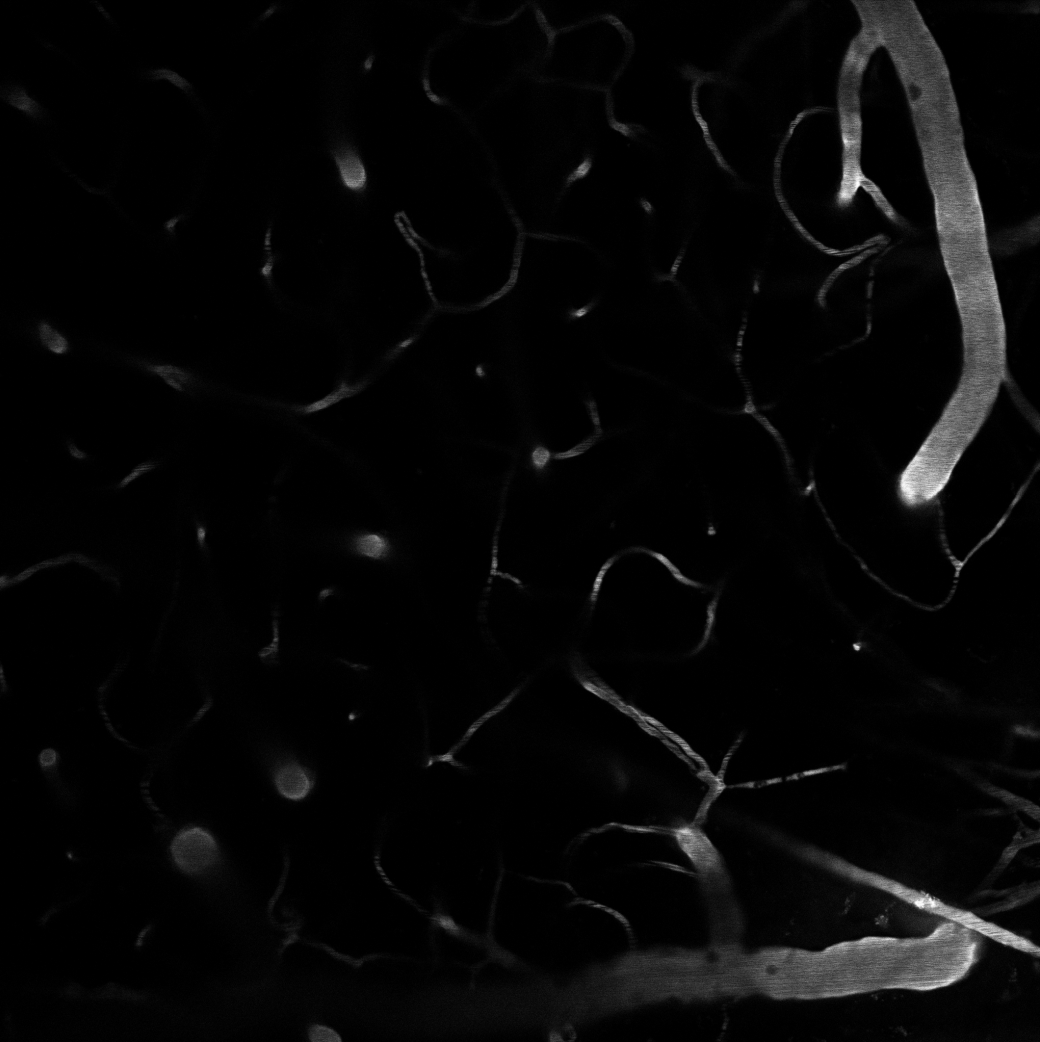}
        \subcaption{Real Blood Vessels}
    \end{minipage}%
    \hfill
    \begin{minipage}{0.45\textwidth}
        \centering
        \begin{minipage}{0.48\textwidth}
            \centering
            \includegraphics[width=\textwidth]{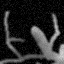}
        \end{minipage}%
        \hfill
        \begin{minipage}{0.48\textwidth}
            \centering
            \includegraphics[width=\textwidth]{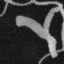}
        \end{minipage}
        
        \vspace{2mm}
        
        \begin{minipage}{0.48\textwidth}
            \centering
            \includegraphics[width=\textwidth]{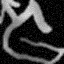}
        \end{minipage}%
        \hfill
        \begin{minipage}{0.48\textwidth}
            \centering
            \includegraphics[width=\textwidth]{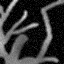}
        \end{minipage}
        \subcaption{Synthetic Homogenous images}
    \end{minipage}

    \vspace{2mm}
    {
    \begin{minipage}{0.21\textwidth}
        \centering
        \includegraphics[width=\textwidth]{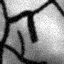}
    \end{minipage}%
    \hfill
    \begin{minipage}{0.21\textwidth}
        \centering
        \includegraphics[width=\textwidth]{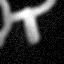}
    \end{minipage}%
    \hfill
    \begin{minipage}{0.21\textwidth}
        \centering
        \includegraphics[width=\textwidth]{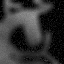}
    \end{minipage}%
    \hfill
    \begin{minipage}{0.21\textwidth}
        \centering
        \includegraphics[width=\textwidth]{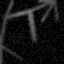}
    \end{minipage}
    \subcaption{Synthetic Varied images}
    }
    \caption{2D projections of real and synthetic 3D blood vessel image scans. The real image on the top left depicts a full scan, which we divide into patches of smaller sizes. The top right contains samples from the Homogenous data, which aims to be as close as possible to the original data patches. The bottom row contains samples from the Varied data, which covers a more diverse set of parameters. These samples include flipped masks, different noise levels, straight vessels, etc.}
    \label{fig:blood_vessel_images}
\end{figure}

Following this pipeline, we create two distinct synthetic datasets: A Homogenous and a Varied dataset whose characteristics are noted in \Cref{tab: datasets}. The Homogeneous dataset comprises 100k samples generated using a single set of parameters carefully tuned to match the characteristics of our real vessel microscopy data. We show several examples from both datasets in \Cref{fig:blood_vessel_images}. In contrast, the varied dataset consists of 1024 samples of 1024 different parameter configurations, yielding roughly 1 million datapoints. This diversity allows for greater variation in vessel network properties, imaging conditions, and artifact patterns. Moreover, this dataset includes configurations that can result in instances that look and are quite different, i.e., by having an inverted mask or consisting of straight connections only. We include a full list of the parameters used when generating the Homogenous and Varied datasets in \Cref{appendix: dataset parameters}. This dual approach enables us to evaluate both the benefits of matched synthetic data and the potential advantages of training on more diverse samples.

\section{Methodology}

We approach the task of vessel network extraction using two separate models that first detect nodes and then predict the connections between them. Our preliminary experiments motivated this separation, which showed that attempting to predict both nodes and edges simultaneously led to reduced accuracy, particularly at complex junctions. By splitting the prediction, we can optimize each stage independently and use ground truth node positions during edge predictor training, eliminating the compounding of localization errors. Both models employ a 3D U-Net based architecture, specifically for the node task we predict both node coordinates and confidence scores. In comparison, the edge prediction stage uses a similar but smaller network that takes the detected node positions as additional input to predict the vessel graph's adjacency matrix directly. Given the typical number of nodes observed in vessel networks from our data, we designed the model to predict a fixed maximum of 32 node coordinates in 3D space along with corresponding confidence scores. The network outputs both a coordinate tensor representing the 3D positions and confidence logits, which are transformed into probabilities using a sigmoid activation.

\begin{figure}[tb]
    \centering
    \includegraphics[width=\linewidth]{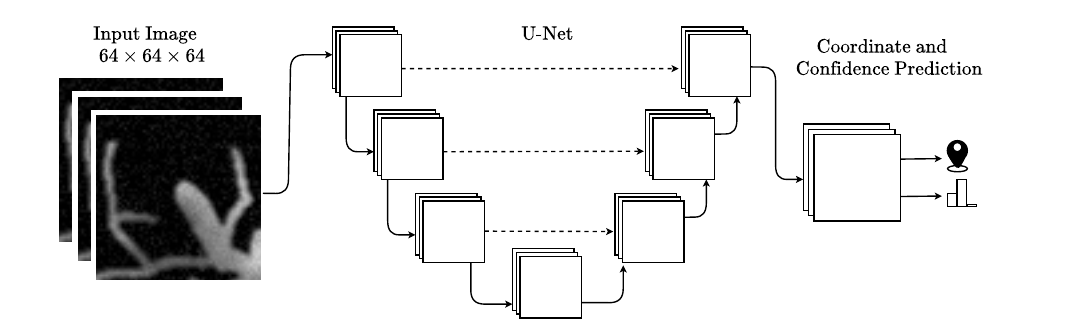}
    \caption{Architecture of the node prediction model, which can detect up to a maximum of 32 nodes. The model follows a U-Net structure with an encoder path and decoder path with additional skip connections between corresponding layers. The final processing block outputs features used by the coordinate prediction head (3D node coordinates) and confidence head (confidence scores).}
    \label{fig: unet_architecture}
\end{figure}


\paragraph{Loss}
The loss function has to balance multiple aspects: a variable amount of present nodes, precise localization as well as confidence calibration. Our approach consists of three components: a coordinate matching loss that ensures precise node localization, an excess prediction penalty that handles predictions without corresponding ground truth nodes, and a confidence loss. The \textit{total loss} is then a combination of these terms:
\begin{equation*}
    \mathcal{L} = \mathcal{L}_{\text{matched}} + \alpha\mathcal{L}_{\text{excess}} + \beta\mathcal{L}_{\text{conf}}
\end{equation*}
where $\alpha,\beta$ are weighting factors between the different losses, $\mathcal{L}_{\text{matched}}$ represents the distance-weighted coordinate matching loss, $\mathcal{L}_{\text{excess}}$ penalizes unmatched predictions based on their confidence and distance to targets, and $\mathcal{L}_{\text{conf}}$ is the binary cross-entropy loss for confidence prediction.

\textit{Coordinate Matching:}
First, we pair the $n$ ground truth nodes with the $n$ most confident predicted nodes using the Hungarian algorithm~\cite{Kuhn1955TheHM}, solving the challenge of arbitrary node ordering. The cost of matching the nodes corresponds to the euchlidian distance between the points. We denote the predictions with $P_i$ and the targets as $T_i$, where $P_i$ was matched with $T_i$ using the Hungarian algorithm. 
Matched predictions contribute to the loss with confidence-based weights:
\begin{equation*}
    \mathcal{L}_{\text{matched}} = \sum_{i < n} \|P_i - T_i\|_2 \cdot (2.0 - \text{conf}_i).
\end{equation*}

\textit{Excess Prediction Penalty:}  
For unmatched predictions, a confidence proportional penalty is applied:
\begin{equation*}
    \mathcal{L}_{\text{excess}} = \sum_{i \geq n} \min_j(\|P_i - T_j\|_2) \cdot \text{conf}_i
\end{equation*} 
This loss ensures that low-confidence unmatched predictions incur a minimal penalty,unmatched predictions with high-confidence are penalized heavily while predictions near targets are penalized less than distant ones.

\textit{Confidence Loss:}
The confidence predictions are additionally trained using a binary cross-entropy loss. For matched predictions, the target confidence is based on the distance to the matched ground truth node, while unmatched predictions are trained towards zero confidence:
\begin{equation*}
    \mathcal{L}_{\text{conf}} = -\sum_{i} \left[t_i \log(\text{conf}_i) + (1-t_i)\log(1-\text{conf}_i)\right]
\end{equation*}
where $t_i = \exp(-d_i)$ is the target confidence for prediction $i$, with $d_i$ being either the distance to the matched target for matched predictions or 1 for unmatched predictions. This formulation ensures that predictions closer to their targets are trained towards higher confidence values.

For edge prediction, we employ a similar U-Net architecture that takes both the image data and detected node positions as input. The network predicts the corresponding adjacency matrix $A$. We ensure symmetry in the adjacency matrix by averaging it with its transpose.
During training, we use the binary cross-entropy loss between the predicted and target adjacency matrix.

\section{Experimental Evaluation}

While our synthetic dataset aims to capture vessel network characteristics as well as possible, the real microscopy z-stack images will always differ. To bridge this gap, we employ a fine-tuning approach using a small set of manually labeled real microscopy data.
Our fine-tuning dataset consists of 10 manually annotated patches of vessel networks. Given its limited size, we split it into five samples for additional training and five for final evaluation. We employ a leave-one-out cross-validation (LOOCV) strategy on the fine-tuning set to maximize insights from the limited data.

We explore two distinct fine-tuning strategies. In the head-only approach, only fine-tunes the final prediction layers while keeping feature extraction layers frozen. This means adjusting only the coordinate prediction and confidence heads for the node prediction model, while for the edge prediction model, only the final edge MLP is trained. In contrast, the full-model approach fine-tunes all layers with a low learning rate $10^{-6}$ to avoid catastrophic forgetting of patterns learned during pre-training.

We use the Homogeneous and Varied datasets to pre-train models. We pretrain the model on the full Varied dataset of roughly one million samples, as it shows continued improvement as more data is used. For the Homogenous data, we tested the performance for the 100k and one million versions of the dataset, where the model trained on the 100k variant performed slightly better. Therefore, this dataset was used to futher fine-tune on the real data. Further details are provided in the Appendix \ref{app:ablation}.  We run both fine-tuning approaches for each pre-trained model by fitting for 100 epochs using the Adam optimizer. After cross-validation identifies optimal hyperparameters, we train the final models using all 5 training samples. This methodology enables us to quantify the benefits of fine-tuning and the relative merits of selective versus full model adjustment when adapting to real microscopy data.

Finally, we also create and evaluate an additional Scanset model, which uses the same architectures but is trained for 1000 epochs exclusively on these 5 training samples to illustrate the impact of using synthetic data for pre-training.

While the synthetic dataset enables training models that can capture basic vessel network characteristics, real microscopy z-stacks present additional challenges and variations not perfectly captured by our synthetic generation pipeline. To bridge this gap, we employ a fine-tuning approach using a small set of manually labeled real microscopy data.

\begin{figure}[t]
    \centering
    \begin{minipage}{0.4\textwidth}
        \centering
        \begin{minipage}{0.37\textwidth}
            \centering
            \includegraphics[width=\textwidth]{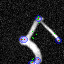}
        \end{minipage}%
        \hspace{0.04\textwidth}
        \begin{minipage}{0.37\textwidth}
            \centering
            \includegraphics[width=\textwidth]{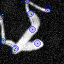}
        \end{minipage}
        
        \vspace{0.04\textwidth}
        
        \begin{minipage}{0.8\textwidth}
            \centering
            \includegraphics[width=\textwidth]{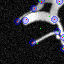}
        \end{minipage}%
        
        \subcaption{Node Prediction}
    \end{minipage}%
    \begin{minipage}{0.4\textwidth}
        \centering
        \begin{minipage}{0.37\textwidth}
            \centering
            \includegraphics[width=\textwidth]{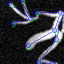}
        \end{minipage}%
        \hspace{0.04\textwidth}
        \begin{minipage}{0.37\textwidth}
            \centering
            \includegraphics[width=\textwidth]{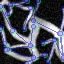}
        \end{minipage}
        
        \vspace{0.04\textwidth}
        
        \begin{minipage}{0.8\textwidth}
            \centering
            \includegraphics[width=\textwidth]{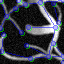}
        \end{minipage}%
        \subcaption{Edge Predictions}
    \end{minipage}

    \caption{2D projections of graph networks predicted on 3D synthetic image data. On the left, we illustrate node prediction. The blue circles indicate the area within a prediction is marked as correct. The red and green dots symbolize the trained model's high and low confidence predictions. }
    \label{fig:predicted}
\end{figure}

\section{Results}

After applying our fine-tuning approaches to node and edge prediction models, we evaluate their performance on the five manually labeled test scans. This directly compares three scenarios: the baseline model without fine-tuning, fine-tuning only the prediction heads, and fine-tuning the entire model. The result is shown in \Cref{fig: fine-tuning performance}.

\begin{figure}[t]
    \centering
    \includegraphics[width=0.95\linewidth]{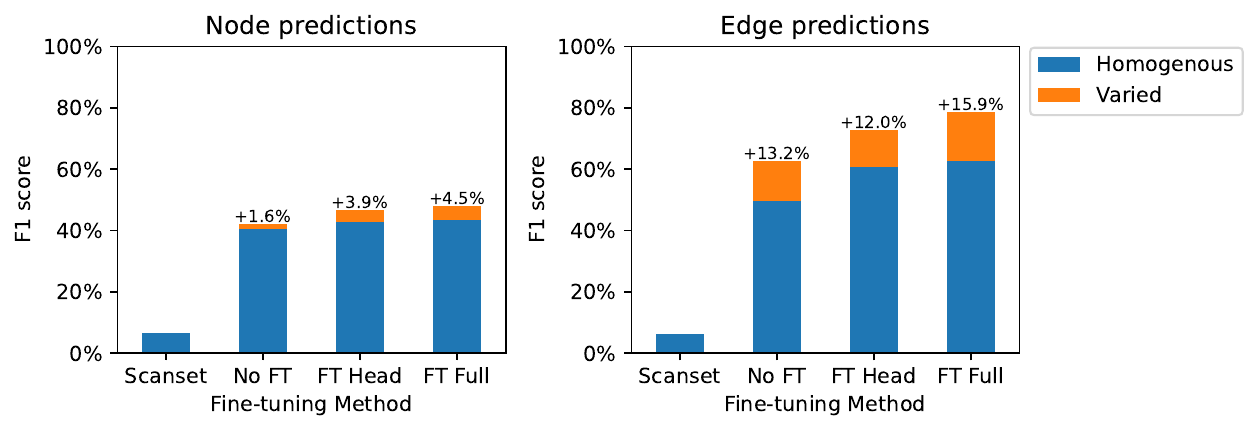}
    \caption{F1 score of node and edge prediction on real-world data with pre-trained models.
    The stacked bars show the improvement of the Varied relative to the Homogenous dataset.
    The values above the bars are the increase of using the Varied dataset. Scanset, only trained on the label real data, does not perform well, and training purely on synthetic data alone improves performance drastically. By fine-tuning the model on very few labeled real data samples, performance can be further improved. Moreover, training on synthetic data that emphasizes diversity in the generation, even if it can result in implausible networks, aids transferring from the synthetic to the real domain. 
    }
    \label{fig: fine-tuning performance no zoom}
\end{figure}

Both fine-tuning approaches show notable improvement over the baseline model for node detection. The Scanset model (F1: 0.067) trained exclusively on the five labeled scans and exhibits very low precision (0.059) and recall (0.077) with a systematic tendency to overpredict, generating 27.2 nodes per image compared to the ground truth average of 20.8. This dramatic performance gap demonstrates that the pre-training on our synthetic data has provided crucial prior knowledge about vessel network structures that cannot be learned effectively from just a handful of real examples.
Edge prediction shows similar improvements with fine-tuning. Interestingly, fine-tuning on the Varied dataset benefits the performance in both the node and edge prediction tasks. This shows that even though the individual data samples might not fully match the blood vessel domain, the additional diversity throughout the dataset leads to better downstream performance. 
The comparison between head-only and full-model fine-tuning reveals only marginal differences, with full-model fine-tuning performing slightly better in both node detection (F1: 0.435 vs. 0.427) and edge prediction (F1: 0.626 vs. 0.607). This suggests that most of the benefits of fine-tuning can be achieved by adjusting only the final prediction layers. 

The results demonstrate that our synthetic data provides a viable path toward vessel network extraction, enabling effective model pre-training that can be refined with very limited real data. The large performance gap between the Scanset and fine-tuned models highlights the value of our approach - while training on just five real samples yields poor results in itself, the same amount of real data can substantially improve a model pre-trained on synthetic data. This suggests that our synthetic data successfully captures key characteristics, allowing the models to learn fundamental features that transfer well to real data. 


\section{Conclusion}

This work presents a novel synthetic data generation pipeline designed to address the challenge of training deep learning models for brain vessel network extraction with very little labeled data. Our parameterized approach generates a diverse and realistic vessel network, enabling the creation of large-scale synthetic training datasets with access to the underlying topological labels. 

We validate the effectiveness of our approach by generating two synthetic datasets and test the performance improvement of a U-Net architecture by first pre-training on the fully synthetic data and then fine-tuning on 5 labeled real data samples. Our results show that pre-training on synthetic data is beneficial, especially on the diverse variation, which improves edge prediction F1 score from 0.496 to 0.626. 
These results demonstrate that our parameterized synthetic data generation approach, combined with strategic fine-tuning, provides a practical solution for the extraction of vessel networks.


\bibliography{iclr2025_conference}
\bibliographystyle{iclr2025_conference}

\appendix

\section{Background and Related Work}
\paragraph{Imaging Techniques}
Understanding blood vessel networks is essential for studying cerebrovascular dynamics and conditions such as stroke, where disruptions in blood flow can lead to severe outcomes. These networks span multiple scales, from large cerebral arteries to fine capillaries, and their analysis requires advanced imaging techniques. 

At the macroscale, magnetic resonance imaging (MRI) and computed tomography (CT) allow non-invasive visualization of major vessels, providing insights into cerebral perfusion and blood flow during neural activation \citep{Ogawa1992}. However, their spatial resolution, typically in the range of hundreds of micrometers, is insufficient for capturing the finer microvasculature.

High-resolution imaging methods, such as two-photon laser scanning microscopy (TPLSM), have become indispensable for studying small vessels and capillaries, offering subcellular resolution \citep{Weber2022, Meng2022}. Through line scanning, TPLSM can quantify absolute blood flow speeds in individual vessels, though it is constrained by small fields of view and the need for invasive cranial windows \citep{Chen2020, Boppart1997}. Recent ultrafast implementations have enhanced temporal resolution, enabling kilohertz frame rates, but remain limited to two-dimensional imaging \citep{Meng2022}. Light microscopy through cranial windows provides a versatile solution for broader fields of view and longitudinal studies, balancing invasiveness with stable optical access to vascular networks \citep{Chen2020}.

\paragraph{Graph Representation}
Graph representations of vascular networks offer a natural framework for analyzing topology and blood flow dynamics. In these representations, vessels are modeled as edges, and junctions serve as nodes, enabling the study of key features such as branching patterns, flow dynamics, and collateral vessel functionality \citep{Blinder2013TheCA, Pia-Flow2024}. This graph-based perspective aligns well with physiological principles, as blood flow entering a junction must equal the flow exiting it. However, extracting these graphs from volumetric microscopy data is fraught with challenges. The data's inherent noise, imaging artifacts, and three-dimensional nature complicate segmentation and graph construction, especially at junction points where errors can disconnect entire regions~\cite{Schmid2017CerebralBF}.

\paragraph{Classic Graph Extraction}
Traditional vessel segmentation and graph extraction approaches typically employ analytical pipelines, as exemplified by \texttt{VesselExpress} \citep{VesselExpress2023}. These methods involve vessel enhancement (e.g., using the Frangi filter), segmentation, centerline extraction, and subsequent graph construction. While interpretable and grounded in established image processing techniques, these methods often fail to maintain topological consistency in complex regions, such as junctions. They are sensitive to noise and parameter variations.

\paragraph{Deep Learning for Graph Extraction}
Deep learning has emerged as a powerful tool for vessel network analysis, shifting the paradigm from multi-stage pipelines to end-to-end solutions. Early advances focused on segmentation tasks, with models like \texttt{U-Net} \citep{UNet2015} and \texttt{CS2-Net} \citep{CS2Net2021} achieving state-of-the-art results across various imaging modalities. However, these models are limited to pixel-level predictions and do not directly output graph representations. More recent methods predict vessel graphs directly, such as \texttt{VesselFormer} \citep{VesselFormer2024} and \texttt{RelationFormer} \citep{Relationformer2022}. \texttt{VesselFormer} extends the transformer-based \texttt{RelationFormer} architecture to predict vessel properties like radius while maintaining global topological relationships. Similarly, frameworks like \texttt{Any2Graph} \citep{Any2Graph2024} employ optimal transport losses, such as the Partially-Masked Fused Gromov-Wasserstein distance, to compare predicted and ground-truth graphs while ensuring permutation invariance. These methods outperform classical pipelines in maintaining topological consistency but require large-scale annotated datasets for training.

\paragraph{Data Challenges}
The scarcity of labeled training data presents a fundamental challenge for both classical and deep learning approaches. Manual annotation of vessel networks is labor-intensive, requiring expert knowledge to ensure accurate topology, especially for 3D datasets \citep{Schmid2017CerebralBF}. Synthetic dataset generation has emerged as a viable solution to this bottleneck, enabling the creation of large-scale datasets with known ground truth. Tools like \texttt{VascuSynth} \citep{vascuSynth2010} simulate vascular growth based on physiological principles such as Murray's law \citep{Murray1926}, vessel curvature constraints, and tissue perfusion requirements. These synthetic datasets provide valuable training data and support the validation of graph extraction algorithms. However, they often fail to capture the full complexity of real vasculature, limiting their direct applicability to real-world datasets.

Efforts to bridge the gap between synthetic and real data include domain adaptation and semi-supervised learning approaches. For example, combining synthetic datasets with limited real-world annotations can improve generalization. Incorporating fluid dynamics considerations \citep{Schmid2017CerebralBF} into synthetic data generation further enhances biological plausibility, creating datasets that more accurately reflect real blood flow patterns. Similarly, work in the area of blood vessel segmentation has looked at taking real data and applying data augmentations designed explicitly for blood vessel data, which can also be a promising way to generate synthetic data \citep{wittmann2024vesselfmfoundationmodeluniversal}. 

Our work focuses on generating synthetic data with a structure and noise pattern similar to those observed in real-world light microscopy images.

\section{Synthetic Dataset Generation}

\subsection{Graph Generation}
Our vessel graph generation follows a growth-based approach inspired by biological vessel formation. Starting from a randomly placed initial vessel, the network grows iteratively, with each new branch maintaining properties observed in real vascular networks. The generation process maintains a priority queue of active vessel segments ordered by their diameter, ensuring that thicker vessels are processed first. As the network expands, new branches are added through single extensions or bifurcations, with vessel diameters scaled to maintain consistent blood flow capacity. The growth continues until either the desired network size is reached or no further valid expansions are possible due to spatial or minimum diameter constraints. This approach naturally creates hierarchical vessel structures that mirror the organization of real vascular networks. 

\subsubsection{Node Placement Strategy}

The foundation of realistic vessel network generation lies in controlled growth rules that create biologically plausible vessel branching patterns. Our algorithm generates nodes through an iterative process that enforces several key constraints:

\begin{itemize}
    \item Creation of branching points with biologically plausible angles and degree distributions
    \item Generation of intermediate extension nodes to create vessel curvature
    \item A minimum distance constraints between nodes to prevent unrealistic clustering
\end{itemize}

This rule-based approach creates graph structures that exhibit characteristics similar to real vessel networks while maintaining computational efficiency and ensuring biological plausibility.

\subsubsection{Edge Connection Rules}
The connection of nodes to form vessel segments is governed by both biological constraints and geometric considerations. Real blood vessels follow specific patterns in how they branch and connect, and our algorithm mimics these patterns through a set of carefully designed rules. 

A key biological constraint we implement is Murray's law \citep{Murray1926}, a law based on the principle of minimum work in biological systems which states that the sum of the cubes of the radii of daughter vessels equals the cube of the radius of the parent vessel:
\begin{equation*}
    r_p^3 = \sum_{i=1}^n r_i^3
\end{equation*}
where $r_p$ is the radius of the parent vessel and $r_i$ are the radii of the daughter vessels.

The branching probability at each vessel segment is calculated based on the current vessel weight (diameter) $w$ and configurable minimum and maximum weights ($w_{min}$, $w_{max}$):
\begin{equation*}
    P_{branch} = 0.2 + 0.6 \cdot \frac{w - w_{min}}{0.8 \cdot w_{max} - w_{min}}.
\end{equation*}
This formula ensures thicker vessels have a higher probability of branching, while very thin vessels rarely branch, matching biological observations.

For each potential branch, spatial validity is verified using intersection detection between line segments. 
Given two vessel segments defined by points $\mathbf{p}_1 = (x_1, y_1)$, $\mathbf{p}_2 = (x_2, y_2)$, $\mathbf{p}_3 = (x_3, y_3)$, and $\mathbf{p}_4 = (x_4, y_4)$, we check for intersection using the cross-product method:
\begin{equation*}
    \begin{aligned}
        d_1 &= (x_4 - x_3)(y_1 - y_3) - (y_4 - y_3)(x_1 - x_3), \\
        d_2 &= (y_4 - y_3)(x_2 - x_1) - (x_4 - x_3)(y_2 - y_1).
    \end{aligned}
\end{equation*}

\subsubsection{Parameter Controls}
The generation process is controlled through a comprehensive set of parameters that ensure flexibility and biological plausibility. The main parameters include:

\begin{itemize}
    \item Vessel diameter range: $w_{min} \leq w \leq w_{max}$
    \item Node spacing constraints: $d_{min} \leq d \leq d_{max}$
    \item Maximum nodes ($N_{max}$) and edges ($E_{max}$)
\end{itemize}

For a generated graph $G = (V, E)$, we want to enforce the following constraints:
\begin{equation*}
    \begin{aligned}
        |V| &\leq N_{max} \\
        |E| &\leq E_{max} \\
        \forall v_i, v_j \in V: &\text{ } d(v_i, v_j) \geq d_{min} \\
        \forall e \in E: &\text{ } d(e.start, e.end) \leq d_{max}
    \end{aligned}
\end{equation*}
where $d(\cdot,\cdot)$ is the Euclidean distance between nodes $v_i$ and $v_j$. These constraints ensure that the generated graphs maintain both biological plausibility and computational feasibility: limiting the total number of nodes and edges prevents excessive complexity, while the distance constraints prevent both unrealistically dense clustering ($d_{min}$) and implausibly long vessel segments ($d_{max}$).

\begin{figure}[t]
    \centering
    \begin{subfigure}{0.3\textwidth}
        \includegraphics[width=\textwidth]{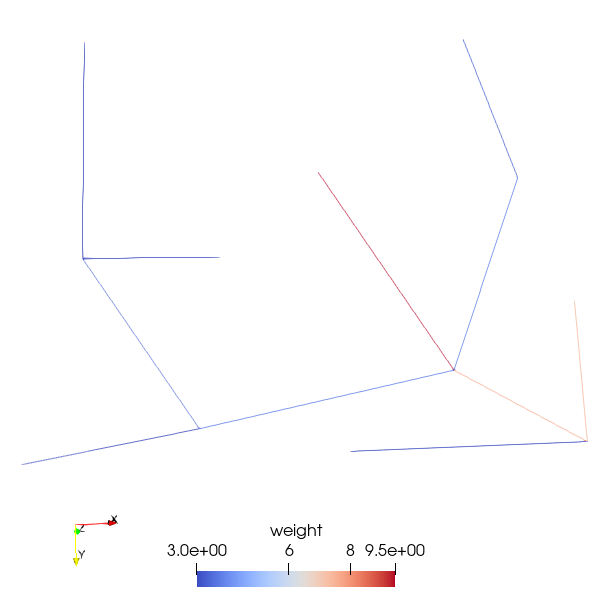}
    \end{subfigure}
    \hfill
    \begin{subfigure}{0.3\textwidth}
        \includegraphics[width=\textwidth]{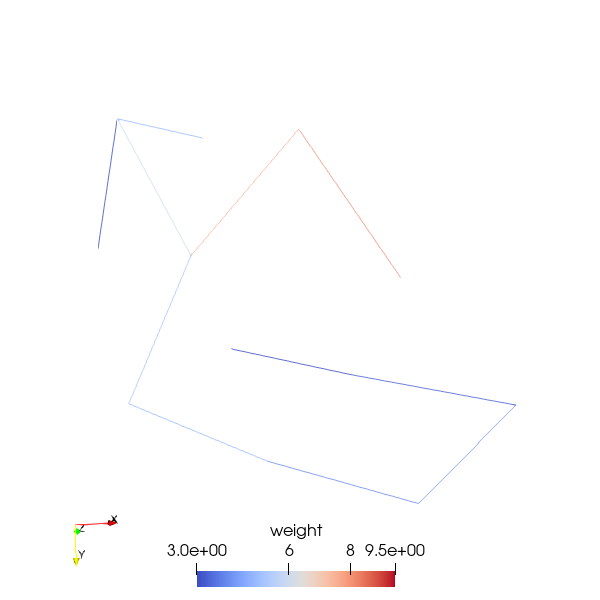}
    \end{subfigure}
    \hfill
    \begin{subfigure}{0.3\textwidth}
        \includegraphics[width=\textwidth]{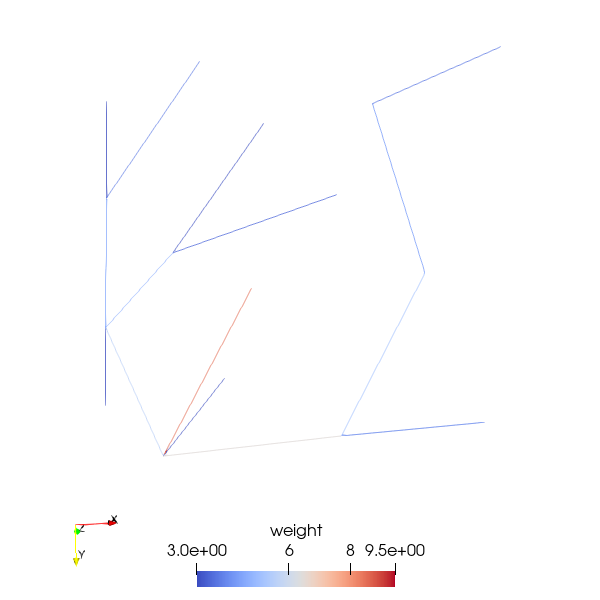}
    \end{subfigure}
    \caption{Generated 2D vessel graphs showing varying complexity and branching patterns. The edge weight indicates the vessel diameter.}
    \label{fig:graph_examples}
\end{figure}

When sampling new node positions, we use a distance-based probability distribution that encourages more uniform spacing between nodes. For each potential new node position $p$, we calculate an acceptance probability based on its distance to the nearest existing node. This probability is highest when the new position is far from existing nodes and decreases exponentially as it gets closer to any existing node:
\begin{equation*}
    P_{accept}(p) = \exp(-\frac{\min_{v \in V} |p - v|^2}{2\sigma^2})
\end{equation*}
where $\sigma = d_{min}/3$ is derived from the minimum distance parameter. This formulation means that positions very close to existing nodes (within $d_{min}$) are rejected, while positions further away are more likely to be accepted. The exponential decay creates a soft boundary rather than a hard cutoff, allowing occasional closer placement while maintaining good average spacing (See \Cref{fig:graph_examples}). This approach helps prevent unrealistic clustering of nodes while allowing natural variations in node density that might occur at vessel bifurcations or in regions of complex vessel structure.

\subsection{Mask Generation}
\subsubsection{Converting Graphs to Binary Masks}
The conversion from graph representation to binary mask involves creating smooth vessel paths and maintaining appropriate vessel thickness. We use cubic Bézier curves for each vessel segment to create natural-looking paths. Given control points $P_0$ through $P_3$, the curve point $B(t)$ at parameter $t$ is:
\begin{equation*}
B(t) = (1-t)^3P_0 + 3(1-t)^2tP_1 + 3(1-t)t^2P_2 + t^3P_3.
\end{equation*}
For each vessel segment with start point $P_0$ and end point $P_3$, we calculate intermediate control points $P_1$ and $P_2$ to create a natural curve:
\begin{equation*}
    \begin{aligned}
        P_1 &= P_0 + \frac{\vec{d}}{3} + w\vec{n} \\[1em]
        P_2 &= P_0 + \frac{2\vec{d}}{3} - w\vec{n}
    \end{aligned}
\end{equation*}
where $\vec{d} = P_3 - P_0$ is the vessel direction vector, $w$ is the vessel weight (diameter), and $\vec{n}$ is the unit normal vector perpendicular to $\vec{d}$. The control points $P_1$ and $P_2$ are positioned one-third and two-thirds along the path respectively, with opposite perpendicular offsets to create a smooth, natural-looking curve.

\begin{figure}[t]
    \centering
    \begin{subfigure}{0.3\textwidth}
        \includegraphics[width=\textwidth]{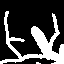}
    \end{subfigure}
    \hfill
    \begin{subfigure}{0.3\textwidth}
        \includegraphics[width=\textwidth]{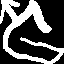}
    \end{subfigure}
    \hfill
    \begin{subfigure}{0.3\textwidth}
        \includegraphics[width=\textwidth]{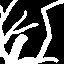}
    \end{subfigure}
    \caption{Binary vessel masks generated from the graph structures. Note how the vessel thickness varies smoothly along branches and at bifurcations.}
    \label{fig:mask_examples}
\end{figure}

\subsubsection{Adding Realistic Vessel Features}
Along these Bézier curves, we place spheres with varying radii to create organic-looking vessels. At each point along the curve, we add subtle random variations to the base radius:
\begin{equation*}
    r_{actual} = r_{base}(1 + \epsilon), \quad \epsilon \sim U(-0.1, 0.1)
\end{equation*}
where $U(-0.1, 0.1)$ represents a uniform random distribution between -0.1 and 0.1. This creates subtle variations in vessel thickness that break up the otherwise perfectly circular cross-section.

The vessel volume is constructed by placing these modified spheres at dense intervals along the Bézier curves. A point $(x, y, z)$ is included in the vessel mask if:
\begin{equation*}
    \sqrt{(x-x_c)^2 + (y-y_c)^2 + (z-z_c)^2} \leq r_{actual}
\end{equation*}
where $(x_c, y_c, z_c)$ is the center of the current sphere. At bifurcations, the overlapping of multiple spheres naturally creates smooth transitions between vessels of different diameters.

\begin{figure}[t]
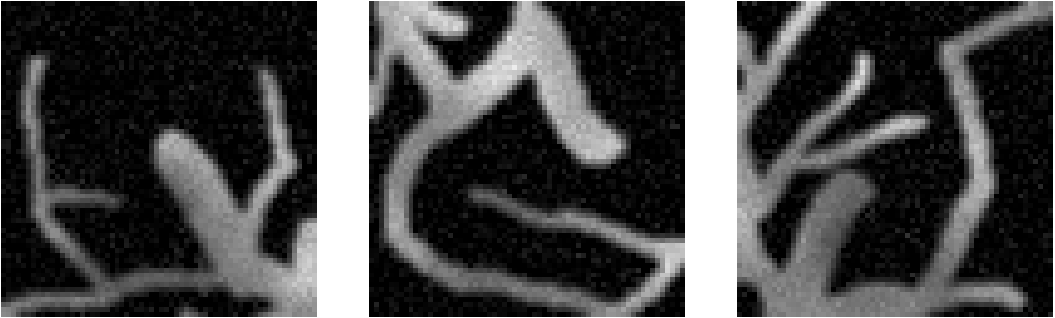

    \centering
    \begin{subfigure}{0.3\textwidth}
        \includegraphics[width=\textwidth]{figures/381/raw.png}
    \end{subfigure}
    \hfill
    \begin{subfigure}{0.3\textwidth}
        \includegraphics[width=\textwidth]{figures/441/raw.png}
    \end{subfigure}
    \hfill
    \begin{subfigure}{0.3\textwidth}
        \includegraphics[width=\textwidth]{figures/655/raw.png}
    \end{subfigure}
    \caption{Final synthetic images.}
    \label{fig:raw_examples}
\end{figure}

\subsection{Adding Noise}
\subsubsection{Adding Realistic Noise and Artifacts}

We add several artifacts to simulate tissue-like structures and natural variations in medical imaging.
A background tissue structure is simulated using Perlin noise \citep{Perlin1985} with multiple octaves:
\begin{equation*}
    P(x,y,z) = \sum_{i=1}^n \frac{1}{2^i} \text{noise}(2^ix, 2^iy, 2^iz)
\end{equation*}
where $n$ is the number of octaves, controlling the level of detail in the noise pattern. Additional brightening is applied along vessel structures using a separate Perlin noise pattern to create variations in vessel intensity, modulated by the original vessel mask.

Furthermore, multiple randomly placed blobs are added with varying sizes and intensities, simulating nearby tissue structures. Each blob's intensity profile follows:
\begin{equation*}
    B(x,y) = \exp\left(-\frac{(x-x_c)^2 + (y-y_c)^2}{2\sigma_b^2}\right) \cdot I
\end{equation*}
where $(x_c, y_c)$ is the blob center, $\sigma_b$ controls the blob size, and $I$ is the intensity.

\subsubsection{Simulating Imaging Conditions}
The final image synthesis accounts for physical imaging effects through several steps.
A point spread function (PSF) simulates optical blur:
\begin{equation*}
    k(x) = \frac{1}{Z}\exp\left(-\frac{x^2}{2\sigma^2}\right)
\end{equation*}
where $Z$ is a normalizing constant. This kernel is applied separably in each dimension for computational efficiency.

Poisson noise simulates quantum noise in the imaging process, with intensity-dependent variance:
\begin{equation*}
    N_{poisson}(x,y,z) \sim \text{Poisson}(\lambda = I(x,y,z) \cdot \text{noise\_level}).
\end{equation*}

Additional Gaussian noise represents electronic noise in the imaging system:
\begin{equation*}
    N_{gaussian} \sim \mathcal{N}(0, \sigma^2_{noise}).
\end{equation*}

These effects are combined to produce the final image:
\begin{equation*}
    I_{final} = \alpha(I_{vessel} * PSF + I_{background}) + N_{poisson} + N_{gaussian}
\end{equation*}
where $*$ denotes convolution and $\alpha$ is a contrast factor.

\section{Synthetic Dataset Parameters}\label{appendix: dataset parameters}

\paragraph{Homogenous Dataset}
The homogenous dataset is generated using the parameters shown in \Cref{tab: homogenous dataset parameters}. 


\begin{table}[ht]
    \centering
    \caption{Dataset Parameters Overview. The table presents the configuration parameters used for generating our synthetic dataset, consisting of 1 million samples. The parameters were carefully chosen to simulate realistic vessel networks while incorporating various imaging artifacts and noise characteristics.}
    \label{dataset-parameters}
    \begin{tabular}{lll}
        \multicolumn{1}{c}{\bf Parameter Category} & \multicolumn{1}{c}{\bf Parameter Name} & \multicolumn{1}{c}{\bf Value} \\
        \toprule
        Volume Dimensions & X dimension & 64 \\
        & Y dimension & 64 \\
        & Z dimension & 64 \\
        \midrule
        Graph Properties & Number of graphs & 1,000,000 \\
        & Maximum number of nodes & 32 \\
        & Maximum number of edges & 0 \\
        & Minimum node distance & 5.0 \\
        & Maximum node distance & 30.0 \\
        & Vessel weight (diameter) range & 3.0 - 10.0 \\
        & Straight lines & False \\
                \midrule

        Imaging Parameters & PSF size & 3 \\
        & PSF sigma & 2.0 \\
        & Noise level & 800 \\
        & Background noise & 0.05 \\
        & Gaussian std & 0.005 \\
        & Background brightness & 0.01 \\
        \midrule

        Perlin Noise & Scale & 20.0 \\
        & Strength & 0.3 \\
        & Darkness & 0.1 \\
        \midrule

        Brightening Effects & Scale & 50.0 \\
        & Strength & 0.8 \\
        \midrule

        Blob Parameters & Maximum number & 5 \\
        & Maximum size & 20 \\
        & Complexity range & [1, 4] \\
        & Elongation range & [1, 3] \\
        & Curvature range & [0, 1] \\
        & Intensity range & [0.01, 0.2] \\
    \end{tabular}
    \label{tab: homogenous dataset parameters}
\end{table}

\paragraph{Varied Dataset}
For the varied configurations, we systematically varied parameters across three aspects of the generation pipeline, with the remaining parameters being the same as those for the homogenous dataset.
\begin{enumerate}
    \item \textbf{Graph Structure Parameters}:
    \begin{itemize}
        \item Minimum node distance ($d_{min}$): 3--10 units
        \item Maximum node distance ($d_{max}$): $d_{min}+5$ to $\min(64, d_{min}+25)$ units
        \item Minimum vessel diameter ($w_{min}$): 2--8 units
        \item Maximum vessel diameter ($w_{max}$): $w_{min}+2$ to $\min(24, w_{min}+16)$ units
    \end{itemize}
    
    \item \textbf{Vessel Mask Parameters}:
    \begin{itemize}
        \item 10\% probability of using straight lines as edges instead of Bézier curves
        \item 10\% probability of inverting the vessel mask
    \end{itemize}
    
    \item \textbf{Image Generation Parameters}:
    \begin{itemize}
        \item Point Spread Function: Size (2--7 units) and sigma (0.5--5.0)
        \item Noise characteristics: Level (200--2000)
        \item Perlin noise: Scale (10--100), strength (0.1--0.8), and darkness (0.05--0.3)
        \item Vessel brightening: Scale (20--100) and strength (0.3--1.2)
        \item Blob artifacts: Count (2--15), size (10--40 units), intensity range (0.01--0.4)
    \end{itemize}
\end{enumerate}

\subsection{Network Extraction}
The complete vessel network extraction combines the node and edge prediction models into a sequential pipeline. This unified approach processes volumetric data through two stages: first detecting vessel junction points, then determining their connectivity to reconstruct the complete vessel network graph.

The pipeline implements a modular design where each model operates independently yet shares a common data representation. When processing an input volume, the node prediction model first identifies potential junction points and assigns confidence scores to each detection. These predictions are filtered using a confidence threshold (typically 0.5), producing a set of high-confidence node coordinates in 3D space.

For training, each model is optimized separately using ground truth data:
\begin{itemize}
    \item The node predictor learns from labeled junction points in the training set
    \item The edge predictor trains on known vessel connections, using ground truth node positions
\end{itemize}

This separation of concerns during training simplifies the optimization process and allows each model to focus on its specific task without interference from potential errors in the other stage. The edge predictor, in particular, benefits from training with ground truth node positions, as it can learn connectivity patterns without having to account for node localization errors.
\subsubsection{Node Prediction}

\begin{figure}[htb]
    \centering
    \includegraphics[width=.8\linewidth]{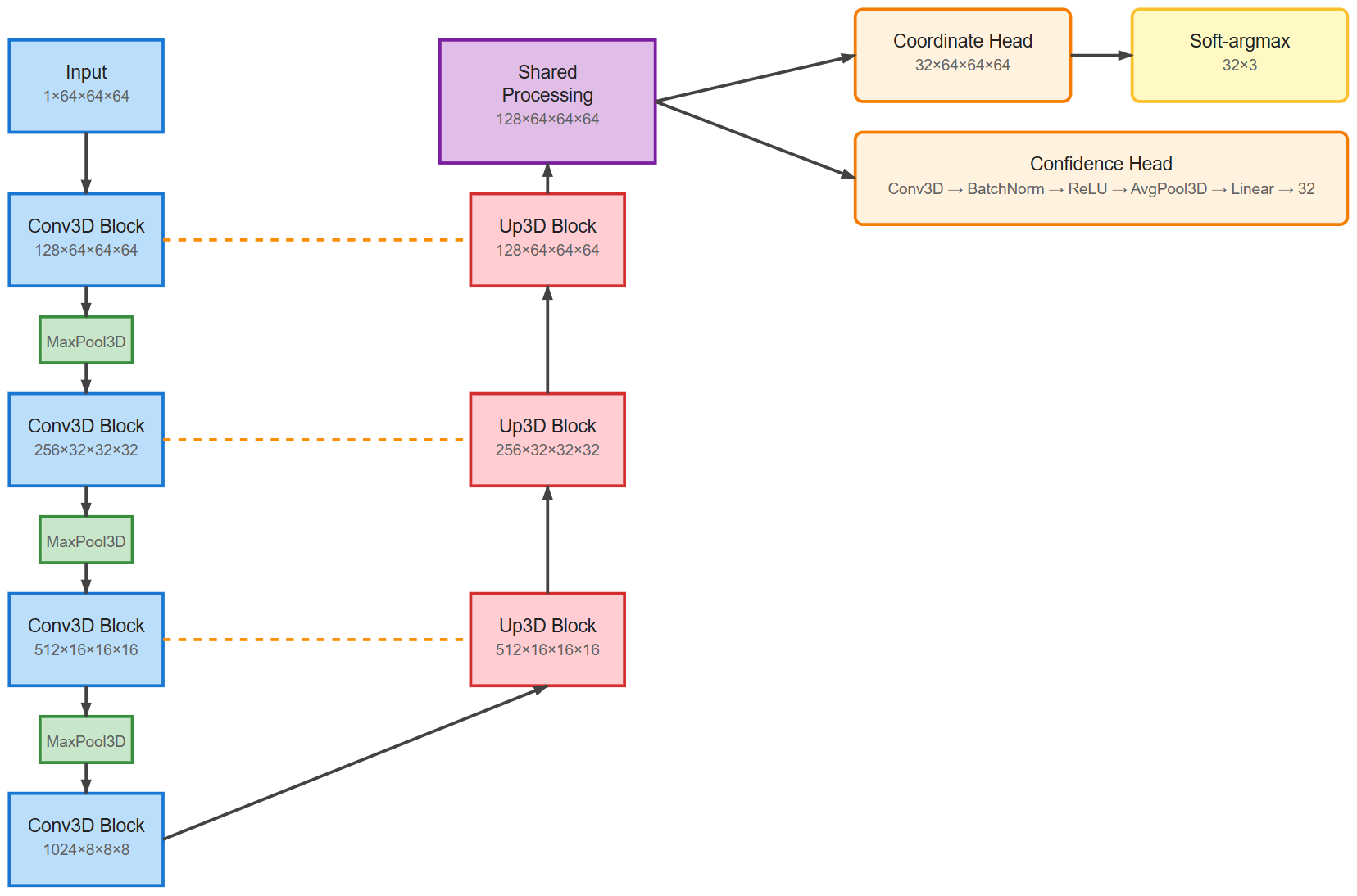}
    \caption{Architecture of the node prediction model for a maximum of 32 nodes. The model follows a U-Net structure with an encoder path (left), decoder path (right), and skip connections between corresponding layers. Numbers indicate tensor dimensions (channels × depth × height × width). The shared processing block outputs features used by both the coordinate prediction head (producing node coordinates in 3D space) and confidence head (producing confidence scores).}
    \label{fig: unet_architecture node}
\end{figure}

\paragraph{Model}
The node detection network (\Cref{fig: unet_architecture node}) builds upon the U-Net architecture~\cite{UNet2015}, extended to handle 3D volumetric data. The network consists of an encoding path that captures context and a decoding path that enables precise localization, with skip connections preserving fine spatial details.

The network predicts both 3D coordinates as well as confidence scores for each potential node. For a maximum of $N$ nodes, the outputs are:
\begin{itemize}
    \item A coordinate tensor $P \in \mathbb{R}^{N \times 3}$ representing the 3D positions
    \item Confidence logits $C \in \mathbb{R}^N$, which are transformed into probabilities using a sigmoid activation
\end{itemize}

This method is inspired by object detection techniques in classical computer vision, where confidence scores are used to filter detections based on their probability of existence.

The confidence score for each node is computed as:
\begin{equation*}
   \text{conf}_i = \sigma(C_i) = \frac{1}{1 + e^{-C_i}}.
\end{equation*}

\subsubsection{Edge Prediction}

\begin{figure}[htb]
    \centering
    \includegraphics[width=.8\linewidth]{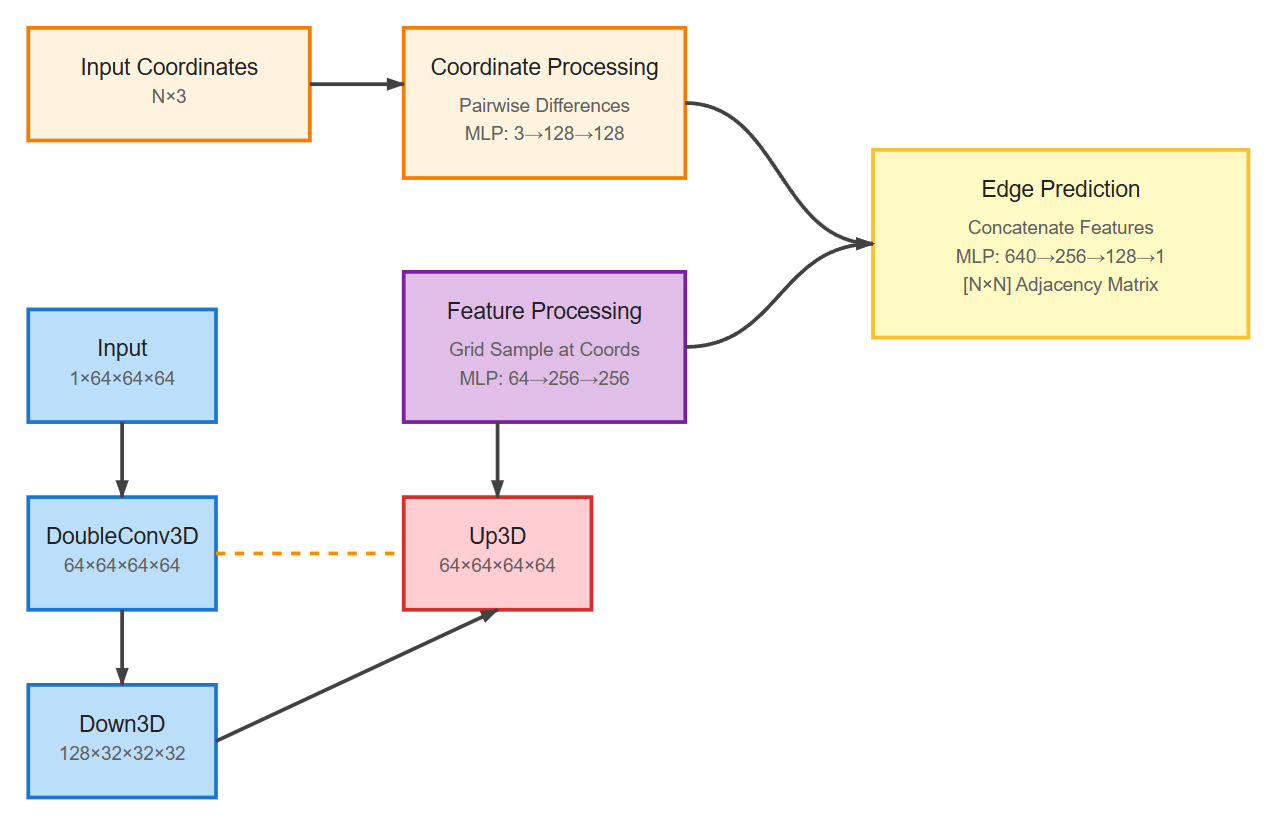}
    \caption{Architecture of the edge prediction model.}
    \label{fig: edge_predictor}
\end{figure}

\paragraph{Model}
The edge prediction network (\Cref{fig: edge_predictor}) addresses the challenge of determining connectivity between detected nodes. Rather than treating this as a pure geometric problem, our approach combines both image features and spatial information. This design choice is motivated by the observation that vessel connectivity is influenced by both the visible vessel structure in the image and the geometric plausibility of connections.

The network processes this information through several stages:
\begin{enumerate}
    \item Image feature extraction using a U-Net-like architecture with:
    \begin{itemize}
        \item Initial feature extraction (64 channels)
        \item Downsampling path doubling features
        \item Upsampling path with skip connections
        \item Final feature map maintaining spatial information
    \end{itemize}
    
    \item Feature extraction at node locations using grid sampling
    \item Feature processing through MLPs to capture node relationships
    \item Final adjacency matrix prediction using node features and coordinate information
\end{enumerate}

The edge probabilities for all node pairs are computed simultaneously through the network, producing an adjacency matrix $A \in [0,1]^{N \times N}$ where $A_{ij}$ represents the probability of an edge between nodes $i$ and $j$. The final adjacency matrix is made symmetric by averaging:
\begin{equation*}
    A_{final} = \frac{A + A^T}{2}.
    \label{eq:symmetic_matrix}
\end{equation*}

\paragraph{Loss}
The edge prediction model takes as input an image and a set of node coordinates, predicting the likelihood of edges existing between each pair of nodes. While during training we typically use ground truth node coordinates, the loss function must also handle the case where node positions come from a previous prediction step. This presents two key challenges:

Firstly, we must establish node correspondence between the input coordinates and the ground truth nodes, as the existence of an edge between nodes $i$ and $j$ only makes sense when we know which nodes these indices refer to. Secondly, vessel networks are undirected graphs, requiring symmetry in the adjacency matrix.

To address the correspondence issue, we apply the Hungarian algorithm to match input nodes with ground truth nodes based on spatial proximity. This produces a reordering matrix $R$ that aligns the nodes with their ground truth counterparts:
\begin{equation*}
    A_{reordered} = RAR^T
\end{equation*}
where $A$ is the predicted adjacency matrix. When using ground truth node coordinates during training, this matching step yields the identity permutation, effectively leaving the adjacency matrix unchanged; the step is nonetheless crucial for correctness when using predicted node positions, ensuring that edges are compared between corresponding node pairs.

The binary cross-entropy loss is then computed between the reordered predictions and the target adjacency matrix:
\begin{equation*}
    \mathcal{L}_{edge} = -\sum_{i,j} [y_{ij}\log(p_{ij}) + (1-y_{ij})\log(1-p_{ij})]
\end{equation*}
where $y_{ij}$ represents the ground truth binary label indicating the existence of an edge between nodes $i$ and $j$, and $p_{ij}$ is the predicted probability of such an edge.

Note that the symmetry requirement for undirected graphs is enforced separately in the model architecture by averaging the prediction matrix with its transpose before applying the sigmoid activation, ensuring that $p_{ij} = p_{ji}$ for all node pairs.

\section{Further Ablations}
\label{app:ablation}
We first validate the impact of the dataset size. To this end, we generate 3 datasets of varying sizes for the Homogenous data, with 10k, 100k, and 1000k samples, respectively. To ensure the performance difference is not caused by the amount of compute, we normalize the compute budgets. Specifically, we train the model with the 1000k-sized dataset for 1 epoch, and then we increase the number of epochs by a factor 10 when we reduce the dataset size by a factor 10. Thus, the models see the same number of samples during training. 

We further test the impact of using the Varied dataset by creating two variations; the first is the full dataset while the second removes 10\% of the parameter classes. We then fit the models on these and test how well they perform on the Homogenous data as that is different from the samples in the Varied dataset.

\begin{figure}[t]
    \centering
    \includegraphics[width=0.8\linewidth]{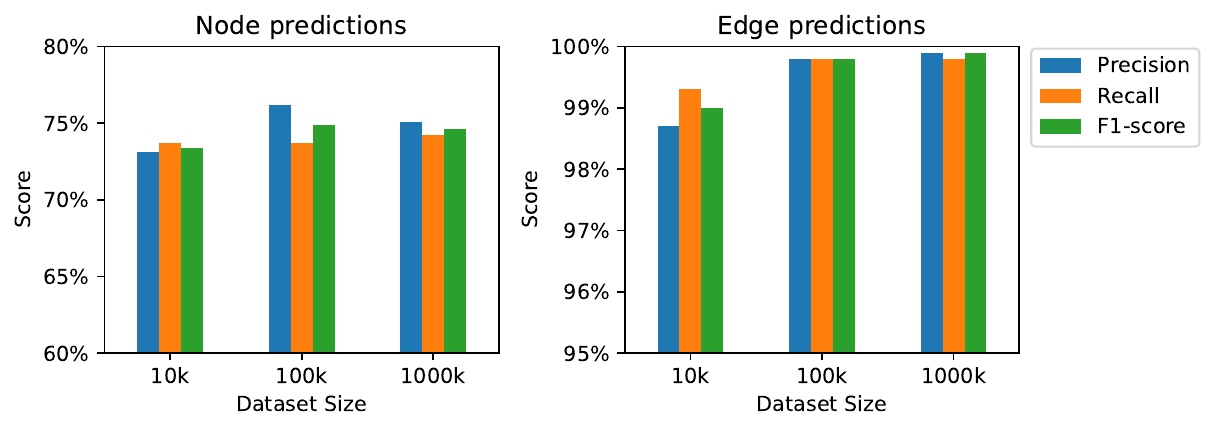}
    \caption{The figure shows the precision, recall, and f1 score for node and edge predictions for different dataset sizes of homogenous data. The left plot shows the results for node predictions, and the right plot shows the results for edge predictions. The x-axis shows the dataset size, and the y-axis shows the score. The y-axis is zoomed in on the relevant regions. See \Cref{fig: dataset size no zoom} for a zoomed-out plot with a y-axis from 0\% to 100\%. 
    The legend shows the color for each score. The figure shows that the precision, recall, and f1 score are somewhat higher for larger dataset sizes, but it levels off between 100k and 1,000k.}
    \label{fig: dataset size}
\end{figure}

The results of the test of dataset size are shown in \Cref{fig: dataset size}. The results show only minor improvements from 10k to 100k and almost no improvements from 100k to 1,000k. Therefore, we deem it unlikely that increasing the size further yields any benefits. Since the precision of node predictions drops from 100k to 1,000k, we use the 100k dataset for later experiments. Note that in most of the subsequent figures, the evalutation is done on the test set of the Homogenous data, where training on the Homogenous data directly seems to be more beneficial. In contrast, performance is better when pretraining on the Varied dataset for the real world data.

\begin{figure}[t]
    \centering
    \includegraphics[width=0.8\linewidth]{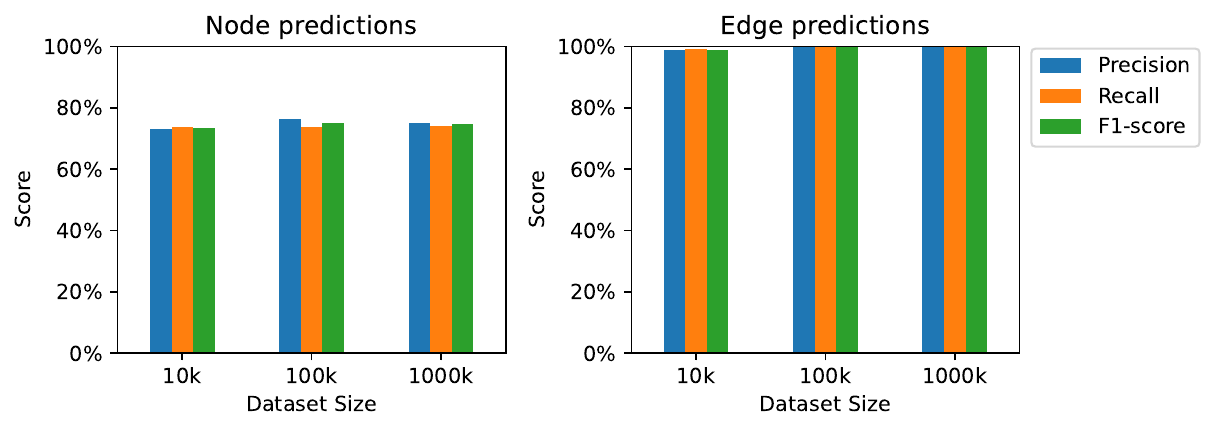}
    \caption{The figure shows the precision, recall, and f1 score for node and edge predictions for different dataset sizes of homogenous data. The left plot shows the results for node predictions, and the right plot shows the results for edge predictions. The x-axis shows the dataset size, and the y-axis shows the score. This is a zoomed-out variant of \Cref{fig: dataset size}. 
    The legend shows the color for each score. The figure shows that the precision, recall, and f1 score are somewhat higher for larger dataset sizes, but it levels off between 100k and 1,000k.}
    \label{fig: dataset size no zoom}
\end{figure}

\begin{figure}[t]
    \centering
    \includegraphics[width=0.6\linewidth]{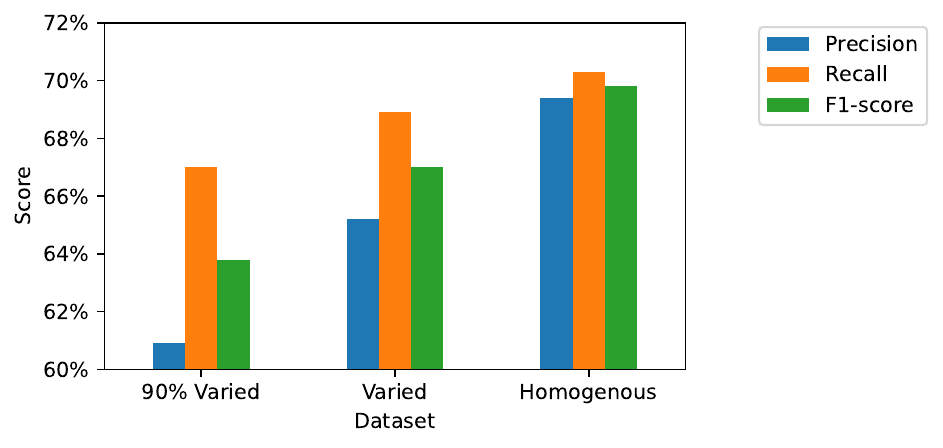}
    \caption{Precision, recall, and f1 score for node predictions on the homogenous test data when training on the 3 datasets, 90\% Varied, Varied, and Homogenous, as indicated by the x-axis. The y-axis indicates the scores, and the view is limited to the relevant region. See \Cref{fig: dataset variety no zoom} for the zoomed-out variant.
    The legend shows the color of each score.
    Pre-training on the homogenous data gives the best performance, but we see a noticeable improvement from 90\% Varied to Varied, indicating that seeing more diverse samples is beneficial.}
    \label{fig: dataset variety}
\end{figure}

\begin{figure}[t]
    \centering
    \includegraphics[width=0.6\linewidth]{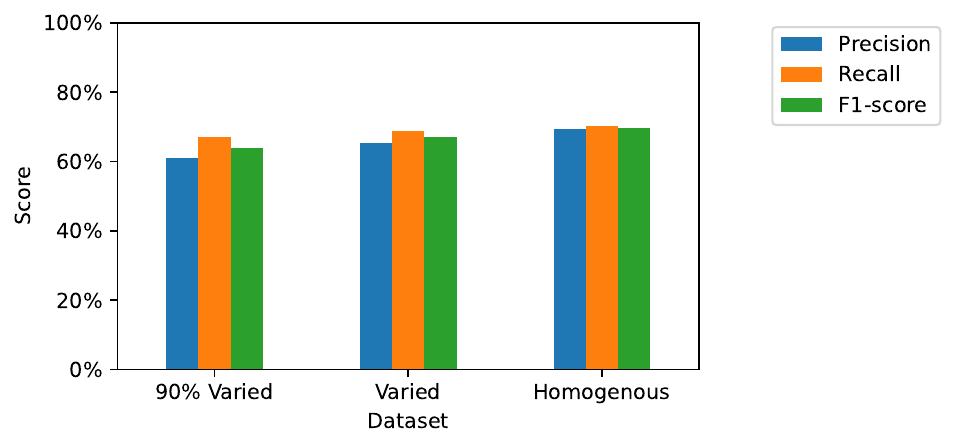}
    \caption{Precision, recall, and f1 score for node predictions on the homogenous test data when training on the 3 datasets, 90\% Varied, Varied, and Homogenous, as indicated by the x-axis. The y-axis indicates the scores, and the view is limited to the relevant region. This is a zoomed-out variant of \Cref{fig: dataset variety}.
    The legend shows the color of each score.
    Pre-training on the homogenous data gives the best performance, but we see a noticeable improvement from 90\% Varied to Varied, indicating that seeing more diverse samples is beneficial.}
    \label{fig: dataset variety no zoom}
\end{figure}

\begin{figure}[t]
    \centering
    \includegraphics[width=1.0\linewidth]{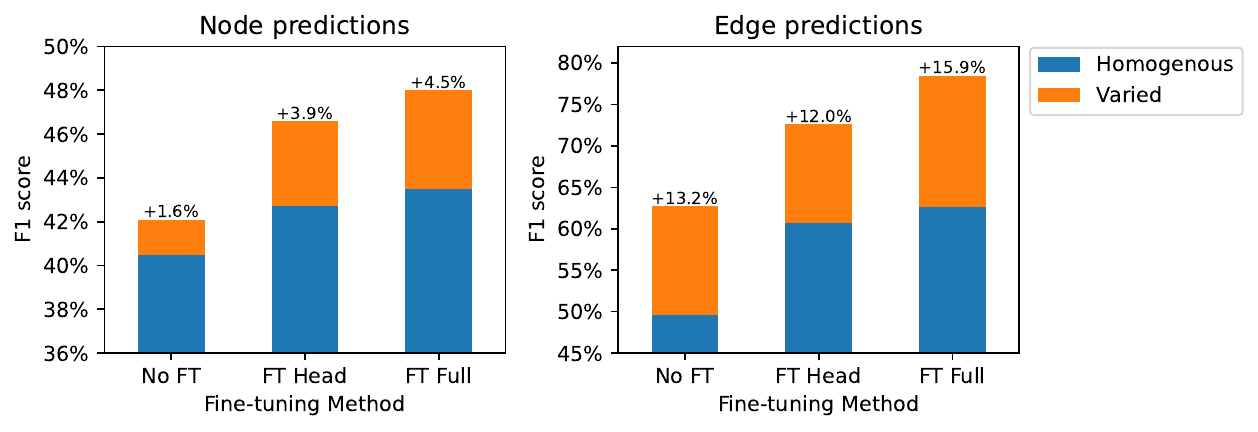}
    \caption{F1 score performance of the node and edge prediction on real-world data with pre-trained models.
    The x-axis indicates the fine-tuning approach, and the y-axis the score. The bars are stacked to show the improvement of the Varied dataset over the Homogenous dataset.
    The values above the bars indicate the increase in \%-points when using the Varied dataset. 
    Without fine-tuning, the f1 score is only 6.7\% and 6.2\% for node and edge predictions, respectively. Thus, we omit it from this zoomed-in plot to make viewing easier. See \Cref{fig: fine-tuning performance no zoom} for the zoomed-out variant with Scanset performance.
    }
    \label{fig: fine-tuning performance}
\end{figure}

\end{document}